\documentclass[letterpaper]{article} % DO NOT CHANGE THIS
\usepackage{aaai25}  % DO NOT CHANGE THIS
\usepackage{times}  % DO NOT CHANGE THIS
\usepackage{helvet}  % DO NOT CHANGE THIS
\usepackage{courier}  % DO NOT CHANGE THIS
\usepackage[hyphens]{url}  % DO NOT CHANGE THIS
\usepackage{graphicx} % DO NOT CHANGE THIS
\usepackage{natbib}  % DO NOT CHANGE THIS AND DO NOT ADD ANY OPTIONS TO IT
\usepackage{caption} % DO NOT CHANGE THIS AND DO NOT ADD ANY OPTIONS TO IT
\usepackage{algorithm}
\usepackage{algorithmic}

\usepackage{newfloat}
\usepackage{listings}

\usepackage{amssymb}
\usepackage{amsmath}
\usepackage{booktabs}
\usepackage{multirow}

\DeclareCaptionStyle{ruled}{labelfont=normalfont,labelsep=colon,strut=off} % DO NOT CHANGE THIS
\floatstyle{ruled}
\newfloat{listing}{tb}{lst}{}
\floatname{listing}{Listing}
\title{Decomposing and Fusing Intra- and Inter-Sensor Spatio-Temporal Signal for Multi-Sensor Wearable Human Activity Recognition}
\author{
    Haoyu Xie\textsuperscript{\rm 1},
    Haoxuan Li\textsuperscript{\rm 2},
    Chunyuan Zheng\textsuperscript{\rm 3},
    Haonan Yuan\textsuperscript{\rm 4},
    Guorui Liao\textsuperscript{\rm 1},
    Jun Liao\textsuperscript{\rm 1},
    Li Liu\textsuperscript{\rm 1,}\thanks{Corresponding author.}
}
\affiliations{
    \textsuperscript{\rm 1}School of Big Data and Software Engineering, Chongqing University, China\\
    \textsuperscript{\rm 2}Center for Data Science, Peking University, China\\
    \textsuperscript{\rm 3}School of Mathematical Sciences, Peking University, China\\    
    \textsuperscript{\rm 4}School of Computer Science and Engineering, Beihang University, China\\
    \{haoyuxie,guoruiliao\}@stu.cqu.edu.cn, \{hxli,cyzheng\}@stu.pku.edu.cn, yuanhn@buaa.edu.cn, \{liaojun,dcsliuli\}@cqu.edu.cn

}

\usepackage{bibentry}
\begin{document}

\maketitle

\begin{abstract}
Wearable Human Activity Recognition (WHAR) is a prominent research area within ubiquitous computing. Multi-sensor synchronous measurement has proven to be more effective for WHAR than using a single sensor. However, existing WHAR methods use shared convolutional kernels for indiscriminate temporal feature extraction across each sensor variable, which fails to effectively capture spatio-temporal relationships of intra-sensor and inter-sensor variables. We propose the \textbf{DecomposeWHAR} model consisting of a decomposition phase and a fusion phase to better model the relationships between modality variables. The decomposition creates high-dimensional representations of each intra-sensor variable through the improved Depth Separable Convolution to capture local temporal features while preserving their unique characteristics. The fusion phase begins by capturing relationships between intra-sensor variables and fusing their features at both the channel and variable levels. Long-range temporal dependencies are modeled using the State Space Model (SSM), and later cross-sensor interactions are dynamically captured through a self-attention mechanism, highlighting inter-sensor spatial correlations. Our model demonstrates superior performance on three widely used WHAR datasets, significantly outperforming state-of-the-art models while maintaining acceptable computational efficiency.
\end{abstract}

%     % \link{Code}{https://github.com/Anakin2555/DecomposeWHAR}
%     % \link{Datasets}{https://aaai.org/example/datasets}
%     % \link{Extended version}{https://aaai.org/example/extended-version}

\section{Introduction}

\begin{figure}[t]
\centering
\includegraphics[width=0.34\textwidth]{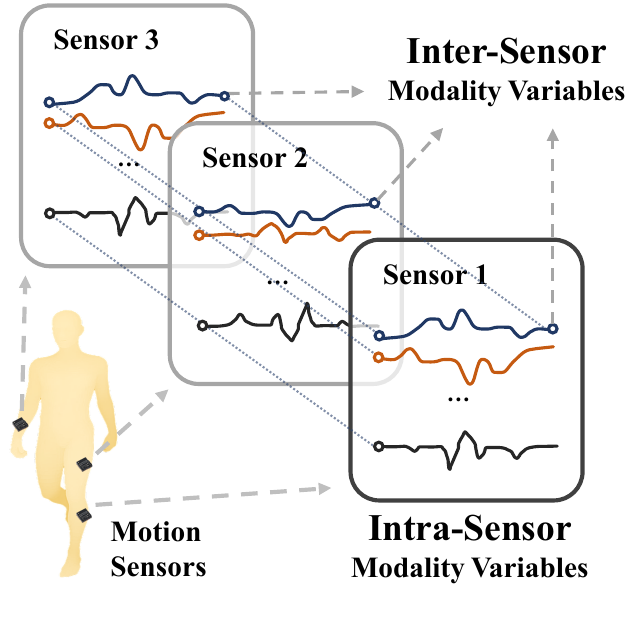} 
\caption{Intra- and Inter-Sensor Variables in WHAR.}
\label{fig:intro}
\end{figure}

Human Activity Recognition (HAR) methods can be categorized into vision-based, environment-based, and wearable sensor-based approaches. Vision-based HAR leverages video data but is restricted by lighting conditions and camera coverage. Environment-based HAR uses ambient sensors such as sound or WiFi, though it faces limitations in location and often struggles with accuracy.

Wearable Human Activity Recognition (WHAR), however, involves attaching sensors directly to the body. This approach offers continuous, direct measurement of motion, is less affected by environmental factors, and is versatile across different settings. It benefits from the use of readily available devices like smartphones and watches, making it practical for applications in healthcare and sports.
WHAR can be classified as a multivariate time series classification problem. The multivariate aspect arises from the various modalities and axes of sensor variables. A typical motion sensor might include a 3-axis accelerometer, a 3-axis gyroscope, and a 3-axis magnetometer, resulting in a 9-variable time series data sample.

% WHAR 中有两种变量关系：intra-sensor和inter-sensor，intra-sensor modality是指同一个sensor内部的不同变量，while inter-sensor modality是指不同传感器之间的变量。但是，过去的方法没能很好地处理好这些关系。
In WHAR, there are two types of variable relationships: \textbf{Intra- and Inter-Sensor Modality Variables} in Figure \ref{fig:intro}. Intra-sensor variables refer to different variables within the same sensor, while inter-sensor variables refer to variables from various sensors located at different positions on the human body. Failing to disentangle these relationships while extracting features will lead to sub-optimal performance.

The first key challenge in WHAR is to capture local and global temporal features of \textbf{Intra-Sensor Modality Variables} while handling the intricate relationships between these diverse variables. Previous studies mainly include two types of approaches. The first type of approach, exemplified by DeepConvLSTM \cite{Ordonez2016Deep} and Attend and Discriminate \cite{Abedin2021Attend}, utilizes shared convolutional kernels to capture local temporal patterns within each variable, followed by Long Short-Term Memory Networks (LSTMs) to integrate variables and model global temporal dependencies. The second type of approach \cite{Miao2022Dynamic, Ahmad2024HyperHAR} employs Conv1D multimodal fusion before temporal extraction, which can lead to the loss of temporal information within each variable. These methods usually pay less attention to retaining the high-level features intrinsic to each variable and modeling complex cross-variable interactions, potentially limiting their ability to fully exploit rich multi-modal intra-sensor data and resulting in reduced recognition performance.

The second challenge is to capture the \textbf{Inter-Sensor} spatio-temporal correlations. For instance, during running, there is a strong correlation between the forward swing of one side's hand and the forward stride of the opposite leg. Similarly, recognizing the relationship between stationary hands and moving legs during cycling can greatly improve activity recognition performance \cite{Ahmad2024HyperHAR}. However, the methods mentioned above typically use dense layers or Recurrent Neural Network (RNN) layers to indiscriminately fuse all variables from all sensors. This simplistic fusion approach fails to capture the spatial relationships between sensors and lacks interpretability. To address this, DynamicWHAR \cite{Miao2022Dynamic} separates sensors placed at different positions and performs multimodal fusion and temporal feature extraction within each sensor individually. It then utilizes a Graph Convolutional Network (GCN) to capture dynamic correlations between sensors. However, GCNs face limitations due to their reliance on predefined graph structures, which restrict their ability to model dynamic relationships and scale effectively with large graphs \cite{yuan2024dynamic}. Another limitation is that GCNs compute correlations between sensors symmetrically, which can overlook the directional and asymmetric nature of interactions between sensors. Additionally, GCNs are less efficient in parallelizing computations due to their message-passing mechanisms \cite{yuan2024environment}.

To overcome these limitations, we introduce a novel framework consisting of a Modality-Aware Signal Decomposition phase and a Hierarchical Interaction Fusion phase. For the first challenge, Modality-Aware Signal Decomposition first separates different sensors to focus on the intra-sensor modality interactions. Inspired by a modern pure convolution structure for time series \cite{Donghao2023ModernTCN}, we leverage a modality-specific embedding layer and local temporal extraction module through Depth-Wise Convolution to independently extract implicit high-level temporal features from intra-sensor modality variable. This ensures both sensor level and variable level independence, capturing detailed temporal features of each modality. 
In the Hierarchical Interaction Fusion phase, we use Point-Wise Convolution to integrate the decomposed channels and modality variables, effectively capturing the intricate dependencies among these intra-sensor modality variables. To address the second challenge, we introduce Global Temporal Aggregation and Cross-Sensor Interaction, implemented by the Mamba-attention module, our model consolidates features across the entire temporal dimension and dynamically captures inter-sensor spatio-temporal correlations. 
Our main contributions are as follows:

\begin{itemize}
    \item We propose a novel framework that incorporates sensor level, temporal level, variable level, and channel level decomposition, followed by reverse-order hierarchical fusion. Our approach effectively captures spatio-temporal relationships between intra- and inter-sensor modality variables, enhancing the recognition of complex patterns in multi-sensor data and diverse scenarios.
    \item To preserve the unique temporal features of intra-sensor modalities, we enhance Depth Separable Convolution to independently extract high-level temporal representations from each modality, mitigating interference between intra-sensor modalities.
    \item We introduce a learnable Mamba-attention module to model inter-sensor correlations dynamically. This module adjusts the relative importance of sensors, addressing heterogeneity across different activities and subjects and enabling more accurate recognition.
    \item Our model achieves superior recognition performance while maintaining computational efficiency, significantly surpassing state-of-the-art baselines.
\end{itemize}

\section{Related Works}

\subsection{Temporal Feature Extraction for WHAR}

Effective temporal feature extraction is essential for WHAR, capturing dynamic patterns from sequential sensor data. Early WHAR methods employed traditional machine learning techniques such as decision trees, SVMs, and k-NN \cite{attal2015physical}. These approaches, limited by feature engineering and temporal modeling, led to the adoption of deep learning methods \cite{Li2021Two-Stream}.
        
Deep learning has significantly enhanced HAR through Convolutional Neural Networks (CNNs). DCNN \cite{yang2015deep} demonstrated CNNs' ability to capture local dependencies, while MCNN \cite{munzner2017cnn} used multi-branch architectures for effective feature fusion. However, CNNs require increased depth to address long-term dependencies, leading to the development of RNNs and their hybrids with CNNs \cite{mutegeki2020cnn, zhou2021automatic}. DeepConvLSTM \cite{Ordonez2016Deep} merges CNNs with LSTMs but faces challenges with long-range dependencies due to the "forgetting" issue.
The self-attention mechanism \cite{vaswani2017attention} has improved long-term dependency modeling. Initial Transformer models for HAR \cite{tonmoy2021hierarchical} and hierarchical encoders \cite{mahmud2020human} effectively capture spatial and temporal features. However, Transformers struggle to correlate isolated timestamps and suffer from quadratic computational complexity with sequence length.

State Space Models (SSMs) have emerged as efficient alternatives. Mamba \cite{dao2024transformers} introduces a selective mechanism and hardware-aware algorithm, offering robust performance across modalities with linear complexity. HARMamba \cite{Li2024HARMamba} further refines this approach for WHAR with bidirectional SSMs. Despite Mamba’s effectiveness in global temporal feature extraction, it shares RNNs' difficulty in managing inter-modality relationships, suggesting that combining Mamba with other models may leverage their complementary strengths.

\subsection{Modeling Intra- and Inter-Sensor Interaction}

Intra-sensor modality fusion techniques have evolved to handle multi-modal data effectively. For instance, some methods \cite{Miao2022Dynamic, Ahmad2024HyperHAR} utilize 1D CNNs to integrate multi-modal information at each timestep. Another approach, Attend and Discriminate \cite{Abedin2021Attend}, employs attention mechanisms following modality-shared convolutional layers to capture cross-modality relationships. However, they overlook the independence of each modality's variables, leading to a potential loss of high-dimensional modal information during fusion.

Regarding inter-sensor interactions, many existing approaches treat variables from all sensors uniformly, which can result in the loss of spatial feature information pertinent to different body parts. Drawing inspiration from STGCN \cite{yan2018spatial} in vision-based human activity recognition, methods like GraphConvLSTM \cite{han2019graphconvlstm}, DynamicWHAR and HyperHAR \cite{Ahmad2024HyperHAR} use GCNs to model spatial correlations between sensors. Despite their advantages, GCNs rely on predefined graph structures, such as human skeleton graphs, which may not fully capture the implicit relationships among sensors.

\section{Preliminaries}

\subsection{Problem Formalization}

Given \( N \) wearable motion sensors, each with \( M \) variables of different modalities (\textit{e.g.}, accelerometer, gyroscope, magnetometer data), the task is to recognize human activities from these multi-sensor data streams. The data can be represented as \(\mathbf{X} = \{ \mathbf{X}_1, \mathbf{X}_2, \ldots, \mathbf{X}_N \} \in \mathbb{R}^{N \times M \times L}\), where \(\mathbf{X}_i \in \mathbb{R}^{M \times L}\) is the data from the \( i \)-th sensor over \( L \) time steps. For example, if each sensor records 3-axis accelerometer (\(a_{x}, a_{y}, a_{z}\)), gyroscope (\(g_{x}, g_{y}, g_{z}\)), and magnetometer (\(m_{x}, m_{y}, m_{z}\)) data, then \(M = 9\).

The objective is to develop a model \( F(\theta) \) that predicts activities \( F(\mathbf{X}) \) from the input \(\mathbf{X}\). The model aims to minimize the difference between the predicted activity \( \hat{\mathbf{y}} = F(\mathbf{X}) \) and the true label \( \mathbf{y} \). The goal is to ensure that the predicted label \( \hat{\mathbf{y}} \) closely matches the actual activity label \( \mathbf{y} \).

\subsection{Depth-Wise and Point-Wise Convolution}

In multi-variable time series data processing, the widely used \textbf{Conv2D} method uses a shared kernel across all channels to learn features spanning multiple variables. While effective for cross-variable correlations, it cannot capture variable-specific patterns.
Inspired by the Depthwise Separable Convolution \cite{chollet2017xception}, \textbf{Depth-Wise Conv1D} applies separate 1D kernels to each channel independently. This method focuses on extracting variable-specific features and offers fewer parameters and faster computation compared to {Conv2D}, thus reducing computational complexity.
Following {Depth-Wise Conv1D}, \textbf{Point-Wise Conv1D} combines information across variables using a kernel size of 1. It performs a linear transformation on the depth dimension, integrating features from depth-wise convolution into a unified representation. 
{Point-Wise Conv1D} further reduces parameter count and enhances computational efficiency, optimizing multi-variable data processing.

\section{Our Model}

\begin{figure*}[t]
\centering
\includegraphics[width=1.0\textwidth]{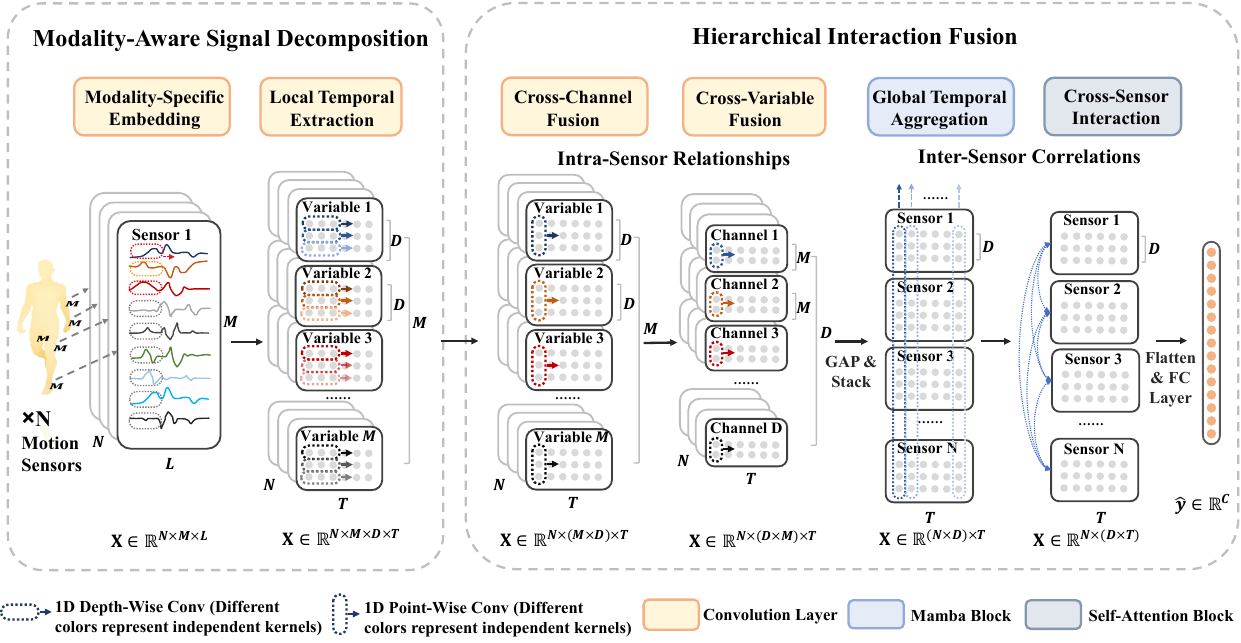} 
\caption{Architecture of Our Model.  GAP represents Global Average Pooling, and FC represents Fully Connected layers.}
\label{fig_model}
\end{figure*}

\subsection{Modality-Aware Signal Decomposition}

We aim to independently extract temporal features from each modality variable within each sensor, free from interference by other variables. This phase ensures both sensor-level and variable-level independence. Initially, each sensor is isolated, and Modality-Specific Embedding (MSE) is performed to convert each sensor’s variables into high-dimensional vectors. Local Temporal Extraction (LTE) is then applied to independently convolve each channel of these high-dimensional vectors, extracting temporal features from each channel. Thus, this phase involves sensor level, variable level, and channel level decomposition, preserving the unique features of each modality.

\subsubsection{Modality-Specific Embedding (MSE).}

We design MSE to transform raw multi-sensor time series data into high-dimensional representations, allowing independent capture of each modality variable's temporal dynamics before further processing.

Given input data from \( N \) wearable motion sensors, each with \( M \) variables and \( L \) time steps, our goal is to embed each variable sequence into a high-dimensional space independently. The input data is represented as a tensor \( \mathbf{X} \in \mathbb{R}^{N \times M \times L} \), where \( \mathbf{X}_n \in \mathbb{R}^{M \times L} \) denotes the data from the \( n \)-th sensor.
The embedding process utilizes independent 1D convolution to transform each variable sequence:
\begin{equation}
\mathbf{X}_{emb} = \text{Conv1D}(\mathbf{X}_n, P, S, D),
\end{equation}
where \( P \) is the kernel size, \( S \) is the stride, and \( D \) is the number of output channels. This operation is performed along the temporal dimension of the variable sequences, resulting in an embedded tensor \( \mathbf{X}_{emb} \in \mathbb{R}^{N \times M \times D \times T} \), where \( T = \frac{L}{S} \) is the new length of the embedded sequences, aiding in reducing the temporal length for more efficient subsequent computations.

The embedding process preserves the unique characteristics of each variable by treating them independently, avoiding interference from other variables.

\subsubsection{Local Temporal Extraction (LTE).}

Local Temporal Extraction captures local temporal features while maintaining the independence of different modality variables. Each variable channel undergoes separate convolution operations, preserving their distinct characteristics.

Given the embedded tensor \( \mathbf{X}_{emb} \in \mathbb{R}^{N \times M \times D \times T} \), where \( N \) is the number of sensors, \( M \) is the number of variables, \( D \) is the number of channels, and \( T \) is the temporal length, we first reshape the tensor to \( \mathbb{R}^{N \times (M \times D) \times T} \). We leverage the depth-wise convolution as Local temporal extraction to capture local temporal dependencies of each variable channel.
The depth-wise convolution is defined as:
\begin{equation}
\mathbf{X}_{dw} = \text{DWConv1D}(\mathbf{X}_{emb}, K_{dw}, G_{dw}= M \times D),
\end{equation}
where \( K_{dw} \) is the kernel size and \( G_{dw} \) is the group number of the depth-wise convolution. This operation is performed separately for each variable channel, ensuring the preservation of unique characteristics. Mathematically, depth-wise convolution can be expressed as:
\begin{equation}
\mathbf{X}_{dw}(n, m, d, t) = \sum_{k=0}^{K_{dw}-1} \mathbf{W}_{m,d,k} \cdot \mathbf{X}_{emb}(n, m, d, t+k),
\end{equation}
where \( \mathbf{W}_{m,d,k} \) denotes the convolutional filter weights for the \( m \)-th variable and \( d \)-th channel.

\subsection{Hierarchical Interaction Fusion}

In the Hierarchical Interaction Fusion phase, we integrate decomposed features at the channel, variable, and sensor levels to capture both intra-sensor and inter-sensor spatio-temporal relationships. This approach begins with combining features within each sensor and then expands to cross-sensor interactions. The process is further refined by the Global Temporal Aggregation module, which consolidates features across the entire temporal dimension, enabling the model to capture long-range dependencies effectively.

\subsubsection{Cross-Channel Fusion (CCF).}

Cross-Channel Fusion merges features across different sensor channels, capturing inter-channel dependencies within each variable.

Starting from the tensor output of depth-wise convolution, \( \mathbb{R}^{N \times (M \times D) \times T} \), point-wise convolutions are performed as follows:
\begin{equation}
\mathbf{X}_{ccf} = \text{PWConv1D}(\mathbf{X}_{dw}, G_{ccf}= M),
\end{equation}
where \( G_{ccf} \) is the group number of the point-wise convolution. This operation fuses information across channels and is followed by reshaping and performing two point-wise convolutions to merge and then restore the dimensionality.

\subsubsection{Cross-Variable Fusion (CVF).}

After Cross-Channel Fusion, the relationships within channels of individual variables are integrated. However, the interactions between different modality variables have not yet been addressed. To capture cross-variable dependencies within the same sensor, we adopt a similar approach to CCF.

The input tensor \( \mathbf{X}_{ccf} \in \mathbb{R}^{N \times (M \times D) \times T} \) is reshaped to \( \mathbf{X}_{ccf} \in \mathbb{R}^{N \times (D \times M) \times T} \), changing the group number to \( D \). Point-wise convolutions are then applied:
\begin{equation}
\mathbf{X}_{cvf} = \text{PWConv1D}(\mathbf{X}_{ccf}, G_{cvf}= D),
\end{equation}
where \( G_{cvf} \) is the group number of the point-wise convolution. This operation fuses information across variables and captures cross-variable dependencies within each sensor.

\subsubsection{Global Temporal Aggregation (GTA).}

The Decomposition Phase extracts modality-specific local temporal features but does not fully capture the temporal context across the entire time series. To address this, we introduce the Global Temporal Aggregation (GTA) module, which consolidates information across all time steps to capture long-range dependencies and overall temporal dynamics.

We start by applying Global Average Pooling (GAP) to reduce each feature map's spatial dimensions to a scalar by averaging over the variable dimension \( M \). For the input tensor \( \mathbf{X}_{cvf} \in \mathbb{R}^{N \times (D \times M) \times T} \), GAP is applied as follows:
\begin{equation}
\mathbf{X}_{gap}(n, d, t) = \frac{1}{M} \sum_{m=1}^{M} \mathbf{X}_{cvf}(n, (d \cdot M + m), t),
\end{equation}
where \( n \) is the sensor index, \( d \) is the channel index, and \( t \) is the time step. This results in \( \mathbf{X}_{gap} \in \mathbb{R}^{N \times D \times T} \), which is then reshaped to \( \mathbf{X}_{stack} \in \mathbb{R}^{(N \times D) \times T} \).

Next, the Mamba block processes \( \mathbf{X}_{stack} \in \mathbb{R}^{(N \times D) \times T} \) to capture critical temporal information. The Mamba block features a Selective SSM that uses linear projections and convolutions to extract local features and selectively retains or discards information. The operation is:
\begin{equation}
\begin{split}
\mathbf{X}_{mb} &= \text{Linear}(\sigma(\text{SSM}(\text{Conv}(\text{Linear}(\mathbf{X}_{stack})))) \\
&\otimes \sigma(\text{Linear}(\mathbf{X}_{stack}))),
\end{split}
\end{equation}
where \( \sigma \) is the activation function and \( \otimes \) denotes element-wise multiplication.
The Mamba block leverages Selective SSM to effectively capture long-term dependencies, ensuring comprehensive temporal feature representation for accurate human activity recognition.

\subsubsection{Cross-Sensor Interaction (CSI).}

To capture inter-sensor spatial correlations, we draw inspiration from the self-attention mechanism, which effectively models relationships between tokens.

Given the output tensor $\mathbf{X}_{mb} \in \mathbb{R}^{(N \times D) \times T}$ from the Mamba block, we reshape it to $\mathbf{X} = \{ \mathbf{X}_1, \mathbf{X}_2, \ldots, \mathbf{X}_N \} \in \mathbb{R}^{N \times (D \times T)}$, where $\mathbf{X}_i \in \mathbb{R}^{D \times T}$ represents the data from the $i$-th sensor. Each $\mathbf{X}_i$ serves as a token for the Attention Layer.
The self-attention mechanism computes responses for each sensor by attending to representations of all sensors. We calculate the normalized correlations across all pairs of sensor data $\mathbf{X}_i$ and $\mathbf{X}_{i'}$ using an embedded Gaussian function. The attention score $\mathbf{A}_{i,i'}$ measures the relevance of data from sensor $i'$ for refining representations of sensor $i$ and is computed as:
\begin{equation}
\mathbf{A}_{i,i'} = \frac{\exp\left({Q}(\mathbf{X}_i)^\top {K}(\mathbf{X}_{i'})\right)}{\sum_{i'=1}^{N} \exp\left({Q}(\mathbf{X}_i)^\top {K}(\mathbf{X}_{i'})\right)},
\end{equation}
where ${Q}$ is the query function projecting the sensor into the query space, and ${K}$ is the key function projecting the sensor into the key space.
These correlations are then used to generate self-attention feature maps $\mathbf{O}_i$ for each sensor:
\begin{equation}
\mathbf{O}_i = {W}\left(\sum_{i'=1}^{N} \left(\mathbf{A}_{i,i'} {V}(\mathbf{X}_{i'})\right)\right),
\end{equation}
where ${W}$ is a linear embedding with learnable weights, and ${V}$ is the value function projecting the sensor into the value space. The feature maps $\mathbf{O}$ are combined with the original sensor data using a residual connection to produce refined feature representations $\mathbf{X}_{csi}$, enabling adaptive integration or exclusion of correlation information.
By employing the CSI module, our model captures interactions between different sensors and encodes these correlations through self-attention weights. During inference, these learned correlations are used to enhance predictions, providing a robust method for synthesizing information from multiple sensors.

\begin{table*}[t]
  \centering
    \resizebox{\textwidth}{!}{
    \begin{tabular}{lcccccccc}
    \toprule
    \multirow{2}[2]{*}{\textbf{Model}} & & \multicolumn{2}{c}{\textbf{Opportunity}} & \multicolumn{2}{c}{\textbf{Realdisp}} & \multicolumn{2}{c}{\textbf{Skoda}} \\
    \cmidrule(r){3-4} \cmidrule(r){5-6} \cmidrule(r){7-8} 
\textbf{} & & {Accuracy (\%)} & {Macro-F1 (\%)} & {Accuracy (\%)} & {Macro-F1 (\%)} & {Accuracy (\%)} & {Macro-F1 (\%)} \\
\midrule
DeepConvLSTM & & 69.30 $\pm$ 0.12 & 61.32 $\pm$ 0.75 & 85.61 $\pm$ 0.63 & 83.56 $\pm$ 0.72 & 90.31 $\pm$ 0.45 & 88.63 $\pm$ 0.39 \\
Att. Model   & & 71.64 $\pm$ 0.53 & 63.43 $\pm$ 0.81 & 88.37 $\pm$ 0.84 & 87.70 $\pm$ 1.06 & 91.55 $\pm$ 0.90 & 90.76 $\pm$ 0.83 \\
GraphConvLSTM & & 70.38 $\pm$ 0.92 & 62.17 $\pm$ 1.05 & 89.94 $\pm$ 0.65 & 89.04 $\pm$ 0.41 & 91.19 $\pm$ 0.68 & 89.88 $\pm$ 0.91 \\
HAR-PBD & & 71.82 $\pm$ 0.67 & 62.37 $\pm$ 0.66 & 91.41 $\pm$ 1.02 & 90.48 $\pm$ 1.14 & 88.37 $\pm$ 1.51 & 86.26 $\pm$ 1.87 \\
Attend and Discriminate & & 72.75 $\pm$ 0.88 & 64.18 $\pm$ 0.43 & 90.46 $\pm$ 0.97 & 89.76 $\pm$ 0.53 & 92.16 $\pm$ 0.98 & 90.87 $\pm$ 0.77 \\
IF-ConvTransformer & & 74.19 $\pm$ 0.94 & 65.92 $\pm$ 0.81 & 90.97 $\pm$ 0.88 & 90.57 $\pm$ 0.62 & 92.27 $\pm$ 0.83 & 90.41 $\pm$ 0.64 \\
DynamicWHAR & & 74.27 $\pm$ 0.46 & {66.13 $\pm$ 0.32} & \underline{92.58 $\pm$ 0.50} & \underline{91.93 $\pm$ 0.57} & {93.80 $\pm$ 0.54} & {91.26 $\pm$ 0.51} \\
HARMamba & & \underline{75.13 $\pm$ 0.81} & \underline{67.05 $\pm$ 0.69} & 91.29 $\pm$ 0.85 & 90.96 $\pm$ 0.73 & \underline{93.96 $\pm$ 0.92} & \underline{91.87 $\pm$ 0.85} \\
(Improvement) & & {1.14\%} & {1.37\%} & {--1.41\%} & {--1.07\%} & {1.75\%} & {0.43\%} \\
\midrule
\textbf{DecomposeWHAR} & & \textbf{78.28 $\pm$ 0.68} & \textbf{72.04 $\pm$ 0.51} & \textbf{96.64 $\pm$ 0.75} & \textbf{96.10 $\pm$ 0.63} & \textbf{97.61 $\pm$ 0.52} & \textbf{97.24 $\pm$ 0.47} \\
\textbf{(Improvement)} & & \textbf{4.02\%} & \textbf{6.93\%} & \textbf{4.21\%} & \textbf{4.34\%} & \textbf{3.74\%} & \textbf{5.52\%} \\
    \bottomrule
    \end{tabular}%
   }
    \caption{Performance comparison of different models under 800 ms window size on the Opportunity, Realdisp, and Skoda datasets. The runner-up’s result is represented with underlined, and the best result is bolded.}
    \label{tab:performance_comparison}%
\end{table*}%

\subsubsection{FC Linear Classifier.}
After the CSI module through the self-attention mechanism, we obtain the output tensor $\mathbf{X}_{csi} \in \mathbb{R}^{N \times (D \times T)}$. This tensor is then reshaped and passed through a Fully Connected (FC) layer for classification:
\begin{equation}
\hat{\mathbf{y}} = \text{FC}(\text{Flatten}(\mathbf{X}_{csi})).
\end{equation}
Here, $\hat{\mathbf{y}} \in \mathbb{R}^{C}$ represents the final output, with $C$ being the number of human activity classes. The predicted activity $\hat{\mathbf{y}}$ is compared to the true label $\mathbf{y}$ using the cross-entropy loss function:
\begin{equation}
\mathcal{L}(\mathbf{y}, \hat{\mathbf{y}}) = - \sum_{i=1}^{C} \mathbf{y}_i \log(\hat{\mathbf{y}}_i),
\end{equation}
where $\mathbf{y}_i$ is the true label, and $\hat{\mathbf{y}}_i$ is the predicted probability for the $i$-th class.

\section{Experiments}

\subsection{Datasets}

To validate the effectiveness and generalizability of our proposed model, we conduct experiments on three widely recognized benchmark datasets in the WHAR community. These datasets are known for their complexity and diversity.

\begin{itemize}
    \item \textbf{Opportunity} \cite{roggen2010collecting}: This dataset involves data from 4 users performing kitchen activities. We focus on data from 5 IMUs placed on the back, right upper arm, right lower arm, left upper arm, and left lower arm, each providing 3-axis accelerometer, gyroscope, and magnetometer at 30 Hz.

    \item \textbf{Realdisp} \cite{banos2012benchmark}: Collected from 17 users performing 33 fitness activities, this dataset uses 9 IMUs on various body parts, including arms, calves, thighs, and back. Each IMU provides data from a 3-axis accelerometer, gyroscope, and magnetometer at 50 Hz. Due to incomplete data, we use recordings from 10 users.

    \item \textbf{Skoda} \cite{stiefmeier2008wearable}: This dataset captures car maintenance activities from one user using 10 accelerometers on both hands, sampled at 98 Hz. 
\end{itemize}

\subsection{Experimental Settings}
Our experiments are conducted using PyTorch on Ubuntu with an RTX 4090 GPU. We optimized the model parameters using the Adam optimizer, setting the learning rate to 0.001 for the Opportunity and Realdisp datasets and 0.0001 for the Skoda dataset. The maximum number of training epochs is set to 80, while setting the batch size to 64 for the Opportunity and Skoda datasets and 128 for the Readisp dataset. The number of attention heads is set to 8. The output dimension ($D$) of MSE is set to 64.

\subsection{Evaluation Protocol}
For the Opportunity and Realdisp datasets, we follow a leave-one-user-out cross-validation approach similar to that used in \citet{chen2022salience, li2021similarity}. In this method, one user is held out as the test subject while the remaining users' data is used for training. This process is repeated until each user has been used as the test subject once, and the average performance across all iterations is reported. While the Skoda dataset contains only one user’s data, by following \citet{Abedin2021Attend, kwon2021approaching}, we employ a hold-out evaluation approach instead of leave-one-subject-out by allocating 80\% of the data to training, 10\% to validation and 10\% to testing, respectively to evaluate our models on the Skoda dataset. The evaluation metrics employed are accuracy and macro F1-score.

\begin{figure}[t]
\centering
\includegraphics[width=0.47\textwidth]{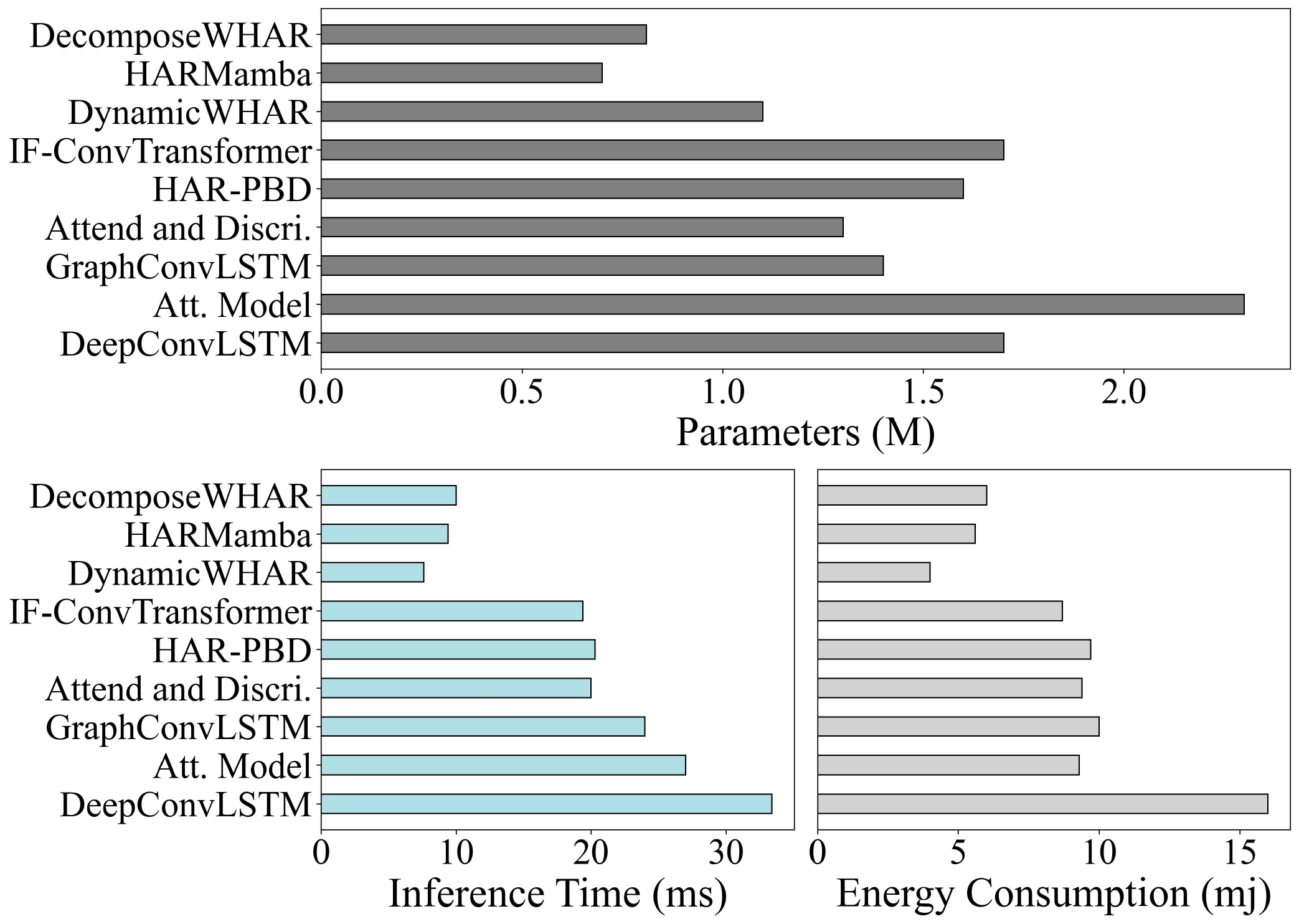} 
\caption{Parameters size and computation efficiency of the Opportunity dataset. The Pytorch model is deployed to the Xiaomi Watch XMWT01 on Wearos 2.41, and inference time and energy consumption are measured. FLOPs are not shown in the figure due to significant discrepancies.}
\label{fig:computation}
\end{figure}

\subsection{Baselines}
{DeepConvLSTM and Its Variants:} {DeepConvLSTM} \cite{Hammerla2016Deep}, {Att. Model} \cite{murahari2018attention}, {Attend and Discriminate} \cite{Abedin2021Attend}. {GCN-based Models:} {GraphConvLSTM} \cite{han2019graphconvlstm}, {HAR-PBD} \cite{wang2021leveraging},  {DynamicWHAR} \cite{Miao2022Dynamic}. {Transformer-based Model:} {IFConvTransformer} \cite{Zhang2022IF-ConvTransformer}. {Mamba-based Model:} {HARMamba} \cite{Li2024HARMamba}.

\begin{table}[t]
  \centering
    \resizebox{0.47\textwidth}{!}{
    \begin{tabular}{lccc}
    \toprule
    \multirow{2}[2]{*}{\textbf{Model}} & \textbf{Opportunity} & \textbf{Realdisp} & \textbf{Skoda} \\
    \cmidrule{2-4}
    & \textbf{Macro-F1(\%)} & \textbf{Macro-F1(\%)} & \textbf{Macro-F1(\%)} \\
    \midrule
    w/o Dec-Phase    &  68.97 ($-$3.07) & 92.03 ($-$4.07) & 93.19 ($-$4.05) \\
    w/o GTA (Mamba)          &  69.33 ($-$2.71) & 90.28 ($-$5.82) & 86.15 ($-$11.09) \\
    w/o CSI (Attention)          &  67.54 ($-$3.50) & 90.85 ($-$5.25) & 79.24 ($-$18.00) \\
    w/o GTA and CSI  &  62.77 ($-$9.27) & 85.54 ($-$10.56)& 77.73 ($-$19.51) \\
    Attention as GTA &  70.26 ($-$1.78) & 95.17 ($-$0.93) & 86.62 ($-$10.62) \\
    GTA after CSI    &  71.40 ($-$0.64) & 94.29 ($-$1.81) & 95.37 ($-$1.87) \\
    \midrule
    \textbf{Full Modules} & \textbf{72.04} & \textbf{96.10} & \textbf{97.24} \\
    \bottomrule
    \end{tabular}}
    \caption{Ablation Study with differences to the best model with full modules enabled (marked in bold).}
  \label{tab:ablation}%
\end{table}%

\subsection{Experimental Results}

\subsubsection{Recognition Performance.}

Table \ref{tab:performance_comparison} compares our method with other state-of-the-art models in terms of accuracy and F1 score. Our method consistently outperforms previous approaches across the Opportunity, Realdisp, and Skoda datasets, showing substantial improvements in both metrics over DynamicWHAR and other earlier models. HARMamaba is implemented by ourselves, and IF-ConvTransformer is based on the released code. The results of other models are quoted from \citet{Miao2022Dynamic}. More details and full results are available in our public repository \footnote{https://github.com/Anakin2555/DecomposeWHAR}.

\subsubsection{Computational Efficiency.}

WHAR applications require accurate recognition while managing computational resources and energy is crucial due to the constraints of wearable embedded devices like smartwatches. Figure \ref{fig:computation} compares model parameters, inference latency, and energy consumption to evaluate efficiency. \textbf{DecomposeWHAR}, HARMamba, and DynamicWHAR have FLOPs under 600 M, whereas others exceed 3000 M. Our model delivers outstanding recognition performance with computational efficiency comparable to the best models, achieved through the Depth Separable Convolution, which reduces parameters and speeds up computation. Additionally, the Mamba block’s selection mechanism, hardware-aware algorithm, and efficient self-attention layer enhance overall efficiency.

\begin{figure}[t]
\centering
\includegraphics[width=0.47\textwidth]{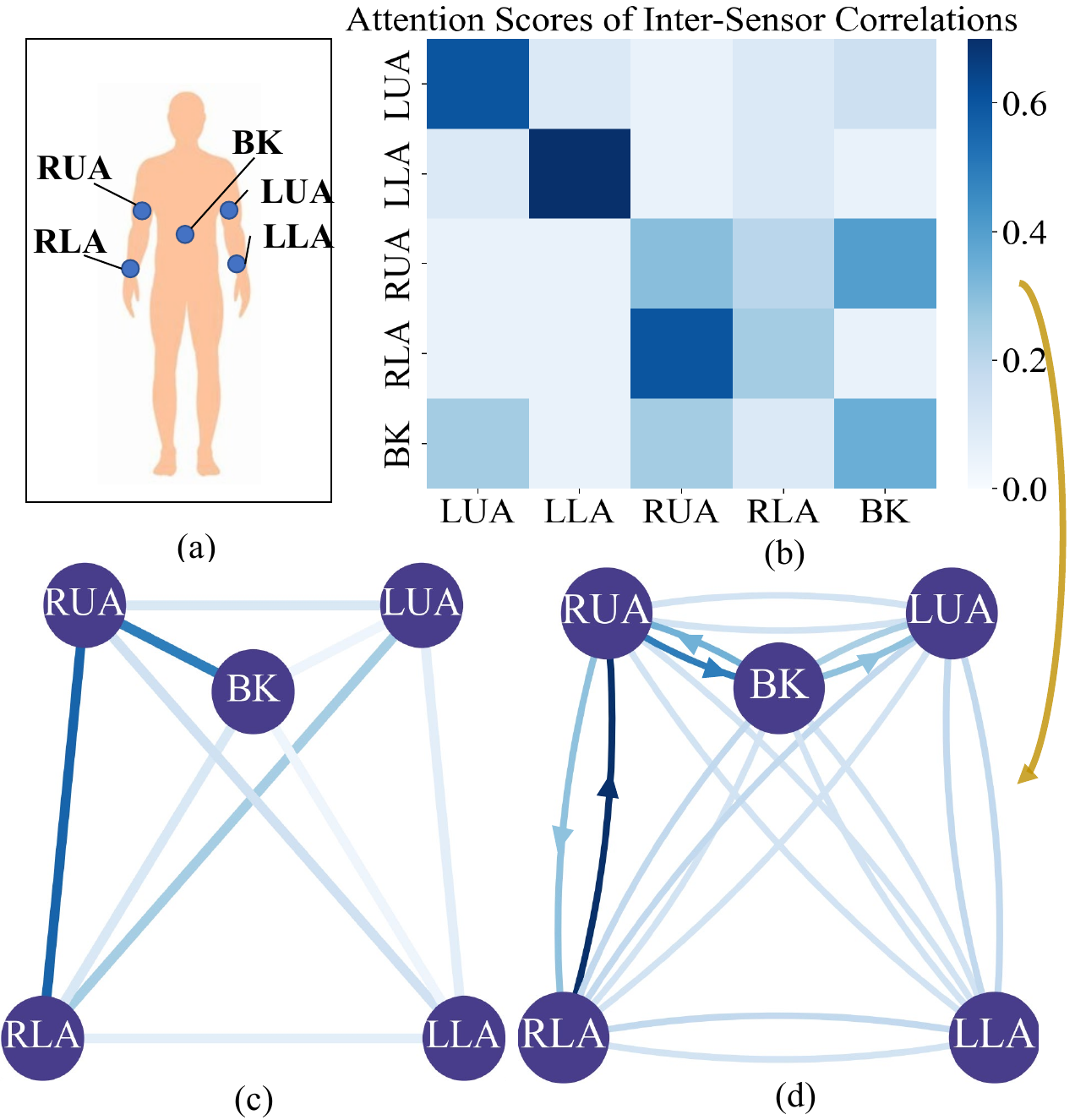} 
\caption{(a) Nodes labelled with "RUA", "RLA", "BK", "LUA", and "LLA" represent the sensors placed on the right up arm, right lower arm, back, left up arm and left low arm. (b) Attention Scores of CSI. (c) Inter-Sensor Correlations of DynamicWHAR. (d) Inter-Sensor Correlations of Ours. The intensity of the line colors represents the strength of the correlations. Only the prominent lines are marked as directed.}
\label{fig:correlation}
\end{figure}

\begin{figure}[t]
\centering
\includegraphics[width=0.47\textwidth]{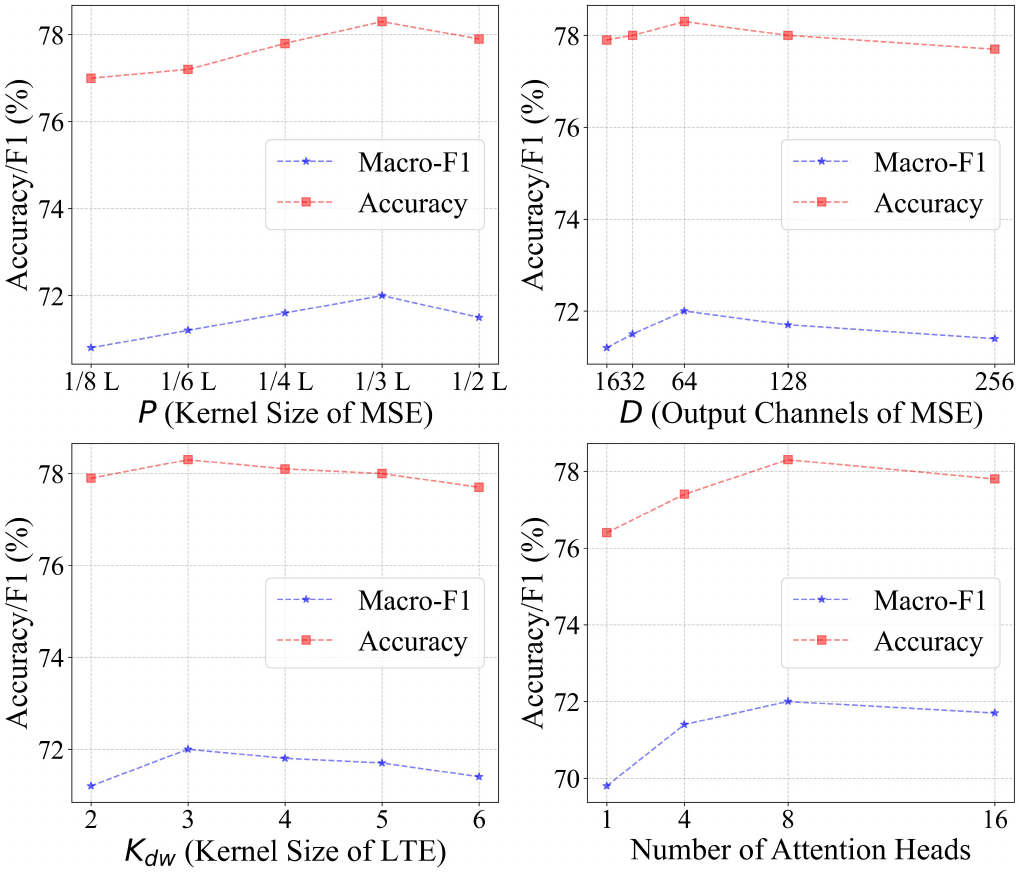} 
\caption{Parameter Analysis.}
\label{fig:parameter}
\end{figure}

\subsection{Ablation Study}

\subsubsection{Effectiveness of Modules.}

In the ablation study, we analyzed the impact of each component on the overall performance in Table \ref{tab:ablation}. Removing the Decomposition Phase (w/o Dec-Phase) and substituting it with Conv2D led to a noticeable decrease in performance. Notably, omitting the Decomposition Phase also eliminates channel and variable fusion. This highlights the importance of the Decomposition Phase in preserving the intra-sensor temporal features. The performance dropped significantly when GTA or CSI was removed, particularly in handling multi-sensor data, demonstrating the critical role of this phase in capturing inter-sensor relationships. Replacing the CSI module with a GCN (GCN as CSI) proposed in DynamicWHAR led to a decline in performance, suggesting that GCNs may not efficiently capture the complex and useful inter-sensor spatial correlations as the CSI module. Using Attention instead of the GTA led to less effective temporal aggregation, showing that GTA is more suited for capturing global temporal patterns. Additionally, reordering the CSI and GTA stages disrupted the intended flow of information processing, reducing the model's ability to integrate sensor interactions effectively.

\subsubsection{Visualization of Inter-Sensor Spatial Correlations.}

To visually demonstrate the effectiveness of capturing inter-sensor spatial correlations, we used the "Drink from Cup" action from the Opportunity dataset, as shown in Figure \ref{fig:correlation}. Our model was compared to the DynamicWHAR model, which uses GCNs to generate sensor interaction strength graphs. DynamicWHAR computes correlations symmetrically through GCNs, potentially missing the directional and asymmetric nature of sensor interactions. For example, during the single-handed drinking action, the attention matrix reveals that the "BK" sensor places slightly more emphasis on "RUA," suggesting that "BK" relies more on "RUA" for maintaining overall posture. Conversely, "RUA" shows relatively less focus on "BK," as it is more concerned with ensuring the accuracy of arm movements rather than the back's posture. Other asymmetrical correlations, such as those between "RUA" and "RLA", also capture more useful inter-sensor spatial information compared to the bidirectional, symmetric interaction strength graphs.

\subsubsection{Parameters Analysis.}

We analyze the impact of different parameter settings on the Opportunity dataset, as illustrated in Figure \ref{fig:parameter}. Similar patterns are observed across other datasets. Our model performs best when $P$ is set to 1/3 of the total time steps ($L$), optimizing the patch size for feature extraction. Setting the Output Channels of MSE ($D$) to 64 effectively retains intrinsic features. A kernel size of 3 for LTE is ideal for capturing local temporal features in WHAR signals and the appropriate number of attention heads is crucial for capturing diverse dependencies and interactions.

\section{Conclusion}

In this paper, we present the \textbf{DecomposeWHAR} model, specifically designed to address the limitations of current WHAR methods by better capturing both intra- and inter-sensor spatio-temporal relationships through Modality-Aware Signal Decomposition and Hierarchical Interaction Fusion. 
In summary, our model shows that decomposing signals from sensor level to channel level and hierarchically fusing them boosts WHAR recognition, offering valuable insights for future research and practical applications.
Our model has potential for further optimization, particularly in reducing parameter size and computational cost. Specifically, the Mamba block in our framework has yet to be refined.
We aim to evaluate the adaptability of our approach to other multivariable classification problems and time series tasks, expanding its applicability across diverse domains.

\section{Acknowledgments}
This work was supported by grants from the National Natural Science Foundation of China (grant Nos. 62477004, 62377040, 62207007, 623B2002) and Chongqing Natural Science Foundation Innovation and Development Joint Fund (grant. no. CSTB2023NSCQ-LZX0109).

\bibliography{aaai25}

\end{document}